\documentclass{article}
\usepackage{spconf,amsmath,graphicx}

\usepackage{booktabs}
\usepackage{multirow}
\usepackage{xcolor}
\usepackage{array}

\usepackage{hyperref}

\usepackage{url}

\usepackage{pifont}

\title{NeXt2Former-CD: Efficient Remote Sensing Change Detection with Modern Vision Architectures}
\name{
Yufan Wang\textsuperscript{1}, 
Sokratis Makrogiannis\textsuperscript{2}, 
Chandra Kambhamettu\textsuperscript{1}
\thanks{This material is based upon work supported by the National Science Foundation under NSF EIR Grant No.~2401835, entitled ``Mapping of Natural Disasters by Deep Subspace Learning in Multi-band and Multi-spectral Satellite Images.''}
}
\address{
\textsuperscript{1}University of South Florida, 
\textsuperscript{2}Delaware State University
}

\begin{document}
%

\maketitle

\begin{abstract}

State Space Models (SSMs) have recently gained traction in remote sensing change detection (CD) for their favorable scaling properties. In this paper, we explore the potential of modern convolutional and attention-based architectures as a competitive alternative. We propose NeXt2Former-CD, an end-to-end framework that integrates a Siamese ConvNeXt encoder initialized with DINOv3 weights, a deformable attention-based temporal fusion module, and a Mask2Former decoder. This design is intended to better tolerate residual co-registration noise and small object-level spatial shifts, as well as semantic ambiguity in bi-temporal imagery. Experiments on LEVIR-CD, WHU-CD, and CDD datasets show that our method achieves the best results among the evaluated methods, improving over recent Mamba-based baselines in both F1 score and IoU. Furthermore, despite a larger parameter count, our model maintains inference latency comparable to SSM-based approaches, suggesting it is practical for high-resolution change detection tasks.
Code will be released at \url{https://github.com/VimsLab/NeXt2Former-CD}

\end{abstract}


\section{Introduction}

Change detection (CD) from bi-temporal remote sensing imagery plays a central role in applications such as urban expansion monitoring and post-disaster assessment. A fundamental difficulty lies in distinguishing true semantic changes from pseudo-changes introduced by illumination variation, seasonal effects, noise, and imperfect co-registration. Early deep learning approaches largely relied on Siamese CNN architectures to extract and compare bi-temporal features. Representative examples include SNUNet-CD~\cite{fang_snunet-cd_2022} and DT-SCN~\cite{liu_building_2021}, which demonstrated strong performance on benchmarks such as LEVIR-CD~\cite{chen_spatial-temporal_2020} and CDD~\cite{lebedev_change_2018}. 
Transformer-based CD methods improve global context modeling by capturing long-range dependencies that are difficult for CNNs with limited receptive fields (e.g., BIT~\cite{chen_remote_2022}, ChangeFormer~\cite{bandara_transformer-based_2022}), but can be computationally expensive on high-resolution imagery.

More recently, State Space Models (SSMs) architectures—in particular, Mamba-style selective scanning—have emerged as an efficient alternative for long-context modeling. Building on evidence in remote sensing representation learning~\cite{chen_rsmamba_2024}, several CD systems adopt Mamba-like backbones and interaction modules, including ChangeMamba~\cite{chen_changemamba_2024}, CDMamba~\cite{zhang_cdmamba_2025}, and M-CD~\cite{paranjape_mamba-based_2024}. At the same time, visual SSMs require serializing 2D features into 1D scan orders, and spatial locality can depend on the chosen traversal. Follow-up studies therefore refine scan strategies or training signals to preserve spatial structure and boundaries~\cite{huang_3d-ssm_2025,wang_change_2025,tobias_samba_2024}. These developments suggest that efficiency-oriented designs may introduce additional choices about how spatial evidence is aggregated during temporal reasoning, and it remains valuable to examine alternative CD pipelines that retain strong 2D inductive biases while achieving competitive accuracy.

\begin{figure*}[t]
  \centering
  \includegraphics[width=\linewidth]{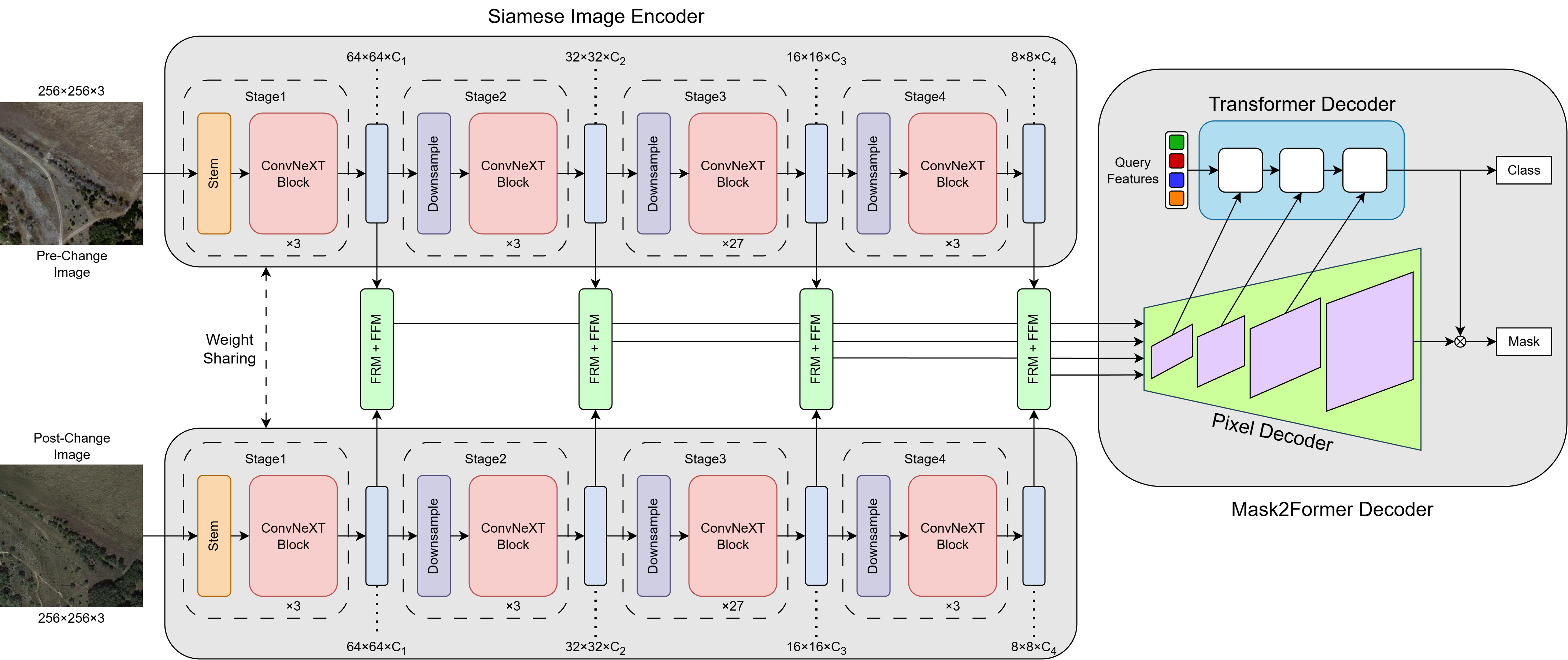}
  \caption{Overview of the proposed change detection network. It consists of a weight-sharing DINOv3~\cite{simeoni_dinov3_2025} backbone, Feature Rectify Modules (FRM) and Feature Fusion Modules (FFM) at multiple scales, and a Mask2Former head for final prediction. Query-level class logits and masks are aggregated into dense predictions via log-sum-exp over queries.}
  \label{fig:architecture}
\end{figure*}

Meanwhile, strong pre-trained backbones and universal decoders have advanced dense prediction: ConvNeXt~\cite{liu_convnet_2022}, Mask2Former~\cite{cheng_masked-attention_2022}, and large-scale self-supervised pre-training such as DINOv3~\cite{simeoni_dinov3_2025}. Building on these components, we propose an end-to-end Siamese CD framework with a DINOv3-pretrained ConvNeXt-L encoder, multi-scale temporal interaction via FRM/FFM~\cite{wan_sigma_2025}, and deformable attention fusion~\cite{zhu_deformable_2021} followed by a Mask2Former head. 
Experiments on LEVIR-CD~\cite{chen_spatial-temporal_2020}, WHU-CD~\cite{ji_fully_2019}, and CDD~\cite{lebedev_change_2018} demonstrate strong performance against recent Mamba-based baselines~\cite{paranjape_mamba-based_2024,chen_changemamba_2024,zhang_cdmamba_2025}.

\section{Related Work}

\subsection{Deep Learning-Based Change Detection}
Deep learning CD is commonly formulated with Siamese encoders and a comparison/decoding head. Representative CNN and attention-based models include SNUNet-CD~\cite{fang_snunet-cd_2022} and DT-SCN~\cite{liu_building_2021}. Transformer-based methods further improve global context modeling, e.g., BIT~\cite{chen_remote_2022} and ChangeFormer~\cite{bandara_transformer-based_2022}. However, the quadratic complexity of self-attention with respect to image resolution can be restrictive for high-resolution remote sensing imagery, motivating exploration of more scalable alternatives. Generative pretraining has also been explored for CD, such as DDPM-CD~\cite{bandara_ddpm-cd_2024}.

\subsection{State Space Models and Mamba in Remote Sensing}
Mamba-style selective scanning has emerged as an efficient alternative to quadratic self-attention for long-context modeling in remote sensing and CD. RSMamba~\cite{chen_rsmamba_2024} demonstrates its potential for remote sensing representation learning, and subsequent CD models include ChangeMamba~\cite{chen_changemamba_2024}, CDMamba~\cite{zhang_cdmamba_2025}, and M-CD~\cite{paranjape_mamba-based_2024}. Many visual SSM designs aggregate context by scanning over 2D feature maps (often in multiple directions) with selective state updates, but maintaining crisp spatial structure and boundary alignment under scan-based interaction remains challenging. Recent work improves spatial coherence via multi-dimensional scanning or additional supervision/augmentation~\cite{huang_3d-ssm_2025,wang_change_2025,tobias_samba_2024}. We explore a complementary direction that keeps temporal interaction explicitly 2D using modern convolutional and attention-based components.

\subsection{Foundation Models and Universal Segmentation}
Pretrained backbones and universal decoders have significantly advanced dense prediction. ConvNeXt~\cite{liu_convnet_2022} and large-scale self-supervised models such as DINOv3~\cite{simeoni_dinov3_2025} provide strong transferable representations, while Mask2Former~\cite{cheng_masked-attention_2022} offers a unified framework for mask prediction. Recent work also explores ConvNeXt+Mask2Former designs in remote sensing dense tasks~\cite{zheng_convnext-mask2former_2024}. Our approach builds on these components and adapts them to Siamese CD with explicit temporal interaction.

\section{Method}

In this section, we present our proposed end-to-end change detection framework. As illustrated in Figure~\ref{fig:architecture}, our method adopts a Siamese encoder-decoder architecture. It leverages a powerful DINOv3-based backbone for feature extraction, a dual-branch feature interaction mechanism to align and fuse temporal information, and a Mask2Former decoder for precise change mask prediction. Unlike recent state-of-the-art methods that rely on state-space models~\cite{paranjape_mamba-based_2024}, feature extraction and decoding in our approach are purely based on convolutional and attention-based mechanisms, achieving superior performance through robust representation learning.

\subsection{Overview}
Given a pair of bi-temporal remote sensing images, the pre-change image $I_1$ and the post-change image $I_2$, our goal is to generate a binary change map $M$ where each pixel indicates the presence or absence of change. The proposed pipeline consists of three main components: (1) A Siamese Backbone that utilizes DINOv3 encoders to extract multi-scale features from $I_1$ and $I_2$; (2) A Feature Interaction and Fusion stage, where features at each scale are processed by a Feature Rectify Module (FRM) and a Feature Fusion Module (FFM) to model spatiotemporal dependencies; and (3) A Change Decoder based on Mask2Former that accepts the fused multi-scale features to predict the final change mask.

\subsection{Siamese DINOv3 Backbone}
To capture robust semantic representations from remote sensing imagery, we employ DINOv3~\cite{simeoni_dinov3_2025} as our backbone encoder. Specifically, we utilize the ConvNeXt-Large variant of DINOv3, which is pre-trained on the LVD-1689M web dataset.

The two input images, $I_1$ and $I_2$, are fed into two parallel branches of the backbone. These branches share identical weights to ensure that the feature extraction process is consistent across time steps. The backbone produces a hierarchy of feature maps at four different scales, corresponding to downsampling strides of 4, 8, 16, and 32 pixels. Let $F_1^i$ and $F_2^i$ denote the feature maps extracted from $I_1$ and $I_2$ at the $i$-th scale, respectively.

\subsection{Spatiotemporal Feature Interaction}
Effective change detection requires not just extracting features but rigorously reasoning about the differences between them while suppressing noise caused by registration errors or seasonal variations. To this end, we incorporate interaction modules inspired by the Feature Rectify and Feature Fusion designs introduced in Sigma~\cite{wan_sigma_2025}.

\noindent\textbf{Feature Rectify Module (FRM).} Before fusion, it is crucial to calibrate the features from one timeframe using information from the other. The Feature Rectify Module (FRM) computes channel and spatial weights based on the concatenation of $F_1^i$ and $F_2^i$. These weights are used to "rectify" the features, effectively highlighting regions of interest and suppressing pseudo-changes. This process is formulated as:
\begin{equation}
  \hat{F}_1^i, \hat{F}_2^i = \text{FRM}(F_1^i, F_2^i).
\end{equation}

\noindent\textbf{Feature Fusion Module (FFM).} To fuse the rectified features, we employ a Feature Fusion Module (FFM). While Sigma~\cite{wan_sigma_2025} adopts a cross-attention-based fusion block, our implementation replaces the cross-attention with deformable attention~\cite{zhu_deformable_2021} to better accommodate geometric deformations and object displacements that can arise in bi-temporal remote sensing pairs. The fusion operation produces a single feature map $Z^i$ for each scale:

\begin{equation}
  Z^i = \text{FFM}(\hat{F}_1^i, \hat{F}_2^i).
\end{equation}
The resulting set of multiscale features $\{Z^i\}$ encapsulates the semantic changes between the two timestamps.

Although LEVIR-CD, WHU-CD, and CDD are broadly co-registered, bi-temporal pairs still exhibit small residual spatial offsets due to orthorectification errors, viewpoint differences, and seasonal effects.
Deformable attention provides adaptive sampling around each spatial location, which is well-suited for handling mild spatial shifts and boundary misalignment in bi-temporal pairs.

\subsection{Mask2Former Decoder and Hybrid Loss}
To translate the fused multi-scale features into a high-quality binary change map, we adapt the Mask2Former~\cite{cheng_masked-attention_2022} decoder. The decoder consists of a pixel decoder that extracts high-resolution per-pixel embeddings from the multi-scale features $\{Z\}$, and a transformer decoder that refines a set of learnable query embeddings via masked attention. This design allows the model to dynamically attend to foreground change regions.

\noindent\textbf{Query-to-Pixel Aggregation.}
Mask2Former~\cite{cheng_masked-attention_2022} predicts a fixed set of $Q$ query outputs, each consisting of a class logit $\mathbf{p}_q$ and a soft mask $\mathbf{m}_q$.
To obtain a dense change map, we aggregate the query-level outputs into pixel-wise logits using a log-sum-exp operation:
\begin{equation}
\ell_c(x,y)=\log\sum_{q=1}^{Q}\exp\!\left(p_q^{c}+m_q(x,y)\right),\quad c\in\{\text{bg},\text{chg}\}.
\label{eq:lse_agg}
\end{equation}
A softmax over the two classes yields the final per-pixel change probability map.

While Mask2Former~\cite{cheng_masked-attention_2022} formulates segmentation as a query-based set prediction problem,
change detection additionally benefits from explicit dense pixel-level supervision.
Our aggregation makes the query-level outputs directly amenable to pixel-wise change reasoning.

\noindent\textbf{Loss Function.} 
Mask2Former~\cite{cheng_masked-attention_2022} formulates segmentation as a query-based set prediction problem and trains the model with Hungarian bipartite matching between predicted queries and ground-truth masks. 
We retain this set-based supervision and additionally introduce an explicit dense pixel-wise loss to improve optimization stability and encourage complete pixel coverage for binary change detection.
 The total loss $\mathcal{L}$ is a weighted combination of the query-based set loss $\mathcal{L}_{set}$ and a dense pixel-wise classification loss $\mathcal{L}_{pixel}$:
\begin{equation}
  \mathcal{L} = \lambda_{set}\mathcal{L}_{set} + \lambda_{pixel}\mathcal{L}_{pixel}.
\end{equation}

The set loss $\mathcal{L}_{set}$ follows Mask2Former~\cite{cheng_masked-attention_2022} and is computed on the matched queries, including a classification loss and mask losses (binary cross-entropy and Dice). 
To complement this query-level supervision, we define a dense pixel-wise loss $\mathcal{L}_{pixel}$ as a weighted cross-entropy over all pixels. 
Concretely, we first convert the $Q$ query outputs into per-pixel class logits via the log-sum-exp aggregation in Eq.~\ref{eq:lse_agg}, and then apply cross-entropy against the binarized ground-truth change mask (no ignore regions). 
We use $\lambda_{pixel}=10$ and $\lambda_{set}=0.1$ in all experiments.

\newcommand{\rr}{\raggedright\arraybackslash}
\begin{table*}[t]
  \centering
  \renewcommand{\arraystretch}{1.15}
  \setlength{\tabcolsep}{4pt}
  \resizebox{\textwidth}{!}{%
  \begin{tabular}{>{\rr}p{4.6cm} c c c c c c c c c c}

    \toprule
    & & \multicolumn{3}{c}{WHU-CD~\cite{ji_fully_2019}} & \multicolumn{3}{c}{LEVIR-CD~\cite{chen_spatial-temporal_2020}} & \multicolumn{3}{c}{CDD~\cite{lebedev_change_2018}} \\
    \cmidrule(lr){3-5}\cmidrule(lr){6-8}\cmidrule(lr){9-11}
    \multirow{-2}{*}{Method} & \multirow{-2}{*}{Extra training data}
      & F1 ($\uparrow$) & IoU ($\uparrow$) & OA ($\uparrow$)
      & F1 ($\uparrow$) & IoU ($\uparrow$) & OA ($\uparrow$)
      & F1 ($\uparrow$) & IoU ($\uparrow$) & OA ($\uparrow$) \\
    \midrule

    \multicolumn{11}{l}{\textcolor{blue}{CNN-based Methods:}}\\
    SNUNet~\cite{fang_snunet-cd_2022} & None
      & 0.835 & 0.717 & 98.7
      & 0.882 & 0.788 & 98.8
      & 0.839 & 0.721 & 96.2 \\
    IFNet~\cite{zhang_deeply_2020} & ImageNet 1k
      & 0.834 & 0.715 & 98.8
      & 0.881 & 0.788 & 98.9
      & 0.840 & 0.719 & 96.0 \\
    \midrule

    \multicolumn{11}{l}{\textcolor{blue}{CNN + Attention based Methods:}}\\
    DT-SCN~\cite{liu_building_2021} & ImageNet 1k
      & 0.914 & 0.842 & 99.3
      & 0.877 & 0.781 & 98.8
      & 0.921 & 0.853 & 98.2 \\
    \midrule

    \multicolumn{11}{l}{\textcolor{blue}{Transformer-based Methods:}}\\
    BIT~\cite{chen_remote_2022} & ImageNet 1k
      & 0.905 & 0.834 & 99.3
      & 0.893 & 0.807 & 98.92
      & 0.889 & 0.800 & 97.5 \\
    ChangeFormer~\cite{bandara_transformer-based_2022} & None
      & 0.886 & 0.795 & 99.1
      & 0.904 & 0.825 & 99.0
      & 0.946 & 0.898 & 98.7 \\
    \midrule

    \multicolumn{11}{l}{\textcolor{blue}{Diffusion-based Methods}}\\
    DDPM-CD~\cite{bandara_ddpm-cd_2024} & Google Earth
      & 0.927 & 0.863 & 99.4
      & 0.909 & 0.833 & 99.1
      & 0.956 & 0.916 & 99.0 \\
    \midrule

    \multicolumn{11}{l}{\textcolor{blue}{Mamba-based Methods}}\\
    RSMamba~\cite{chen_rsmamba_2024} & ImageNet 1k
      & 0.927 & 0.865 & 99.4
      & 0.897 & 0.814 & 98.9
      & 0.943 & 0.902 & 98.8 \\
    ChangeMamba~\cite{chen_changemamba_2024}& ImageNet 1k
      & 0.925 & 0.861 & 99.4
      & 0.902 & 0.821 & 99.0
      & 0.944 & 0.920 & 99.0 \\
    CDMamba~\cite{zhang_cdmamba_2025} & ImageNet 1k
      & 0.937 & 0.882 & 99.5
      & 0.907 & 0.831 & 99.0
      & 0.960 & 0.919 & 99.1 \\
    M-CD~\cite{paranjape_mamba-based_2024} & ImageNet 1k
      & \underline{0.954} & \underline{0.911} & \underline{99.635}
      
      & \underline{0.919} & \underline{0.850} & \underline{99.185}
      
      & \underline{0.981} & \underline{0.963} & \underline{99.540} \\
    \midrule

    NeXt2Former-CD (Ours) & LVD-1689M


      

    & \textbf{0.955} & \textbf{0.914} & \textbf{99.646}

    & \textbf{0.921} & \textbf{0.854} & \textbf{99.202}
    
    & \textbf{0.984} & \textbf{0.969} & \textbf{99.610} \\

      
    \bottomrule
  \end{tabular}}
  \caption{
    Comparison of NeXt2Former-CD with state-of-the-art change detection methods.
    F1 and IoU denote the F1 score and Intersection-over-Union for the change class, and OA denotes overall accuracy.
    For M-CD~\cite{paranjape_mamba-based_2024}, we re-evaluated the officially released models and report the corrected F1 scores (IoU/OA match the M-CD paper; see supplementary material).
    The best result is indicated in \textbf{bold}, and the second-best result is \underline{underlined}.
    }

  \label{tab:mcd_sota_comparison}
\end{table*}

\begin{table}
  \centering
  \renewcommand{\arraystretch}{1.15}
  \setlength{\tabcolsep}{4pt}
  \begin{tabular}{p{0.40\columnwidth} c c c}
    \toprule
    Fusion Strategy 
    & F1 $\uparrow$ 
    & IoU $\uparrow$ 
    & OA $\uparrow$ \\
    \midrule
    Cross-Attn 
      & 0.919 
      & 0.8498 
      & 99.321 \\
    \midrule
    \textbf{Deformable-Attn} 
      & \textbf{0.921} 
      & \textbf{0.8531} 
      & \textbf{99.330} \\
    \bottomrule
  \end{tabular}
    \caption{Ablation study on the use of deformable attention in the Feature Fusion Module (FFM) on LEVIR-CD~\cite{chen_spatial-temporal_2020} with batch size 64, evaluated on the validation split.}
  \label{tab:ablation_deform_attn}
\end{table}

\begin{figure}[t]
  \centering
  \includegraphics[width=\columnwidth]{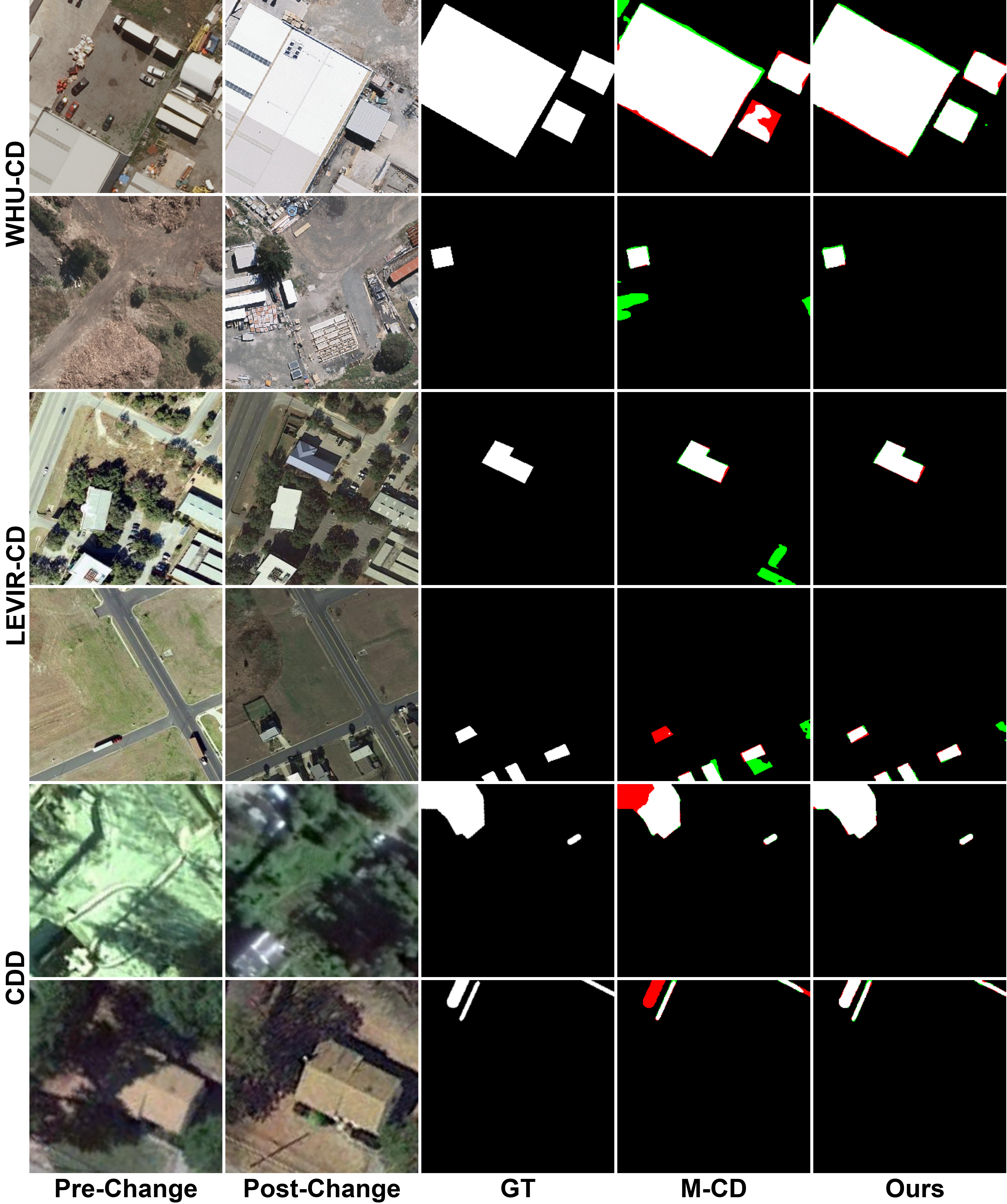}
  \caption{Qualitative results on three public datasets. White represents true positives, black represents true negatives, green represents false positives and red represents false negatives.}
  \label{fig:qualitative}
\end{figure}

\section{Experiments}

\subsection{Datasets}

We evaluate on three public CD benchmarks: LEVIR-CD~\cite{chen_spatial-temporal_2020}, WHU-CD~\cite{ji_fully_2019}, and CDD~\cite{lebedev_change_2018}. Following M-CD~\cite{paranjape_mamba-based_2024}, we use their released pre-processing and splits, where the original images are cropped into non-overlapping $256\times256$ patches. We report F1/IoU for the change class and overall accuracy (OA) on the test set for fair comparison. We exclude DSIFN-CD~\cite{zhang_deeply_2020} because the released M-CD training pipeline becomes numerically unstable on this dataset under the authors' default settings (NaN loss), and we keep all experiments aligned with that protocol for fair comparison.


\subsection{Experimental Setup}

Experiments are conducted on RTX A6000 and RTX 5090 GPUs. We follow the training and evaluation protocol of M-CD~\cite{paranjape_mamba-based_2024}. Models are trained for 150 epochs using AdamW. The encoder is initialized with DINOv3-pretrained ConvNeXt-L weights trained on LVD-1689M~\cite{simeoni_dinov3_2025}. A reduced learning rate (0.1×) is applied to the encoder throughout training, and the encoder is optionally frozen for the first 5 epochs. Performance is measured using F1, IoU for the change class, and overall accuracy (OA).

For metric consistency, we re-evaluate M-CD~\cite{paranjape_mamba-based_2024} using the released checkpoints. We confirm that the resulting IoU and OA match the M-CD paper, while we report the corrected F1; details are provided in the supplementary material.
On CDD, training with batch size 8 on a single RTX 5090 GPU takes 13.39 hours for M-CD and 19.58 hours for our model, with similar peak VRAM usage (18.16 vs. 18.57 GiB).

\begin{table}[b]
  \centering
  \renewcommand{\arraystretch}{1.15}
  \setlength{\tabcolsep}{4pt}

  \begin{tabular}{l c c c}
    \toprule
    Loss Configuration 
    & F1 $\uparrow$ 
    & IoU $\uparrow$ 
    & OA $\uparrow$ \\
    \midrule
    Cross Entropy Loss
      & 0.916 & 0.8456 & 99.294 \\
    Mask2Former Set Loss
      & 0.917 & 0.8474 & 99.316 \\
    \midrule
     \textbf{Hybrid Loss (ours)}
      & \textbf{0.921} & \textbf{0.8531} & \textbf{99.330} \\
    \bottomrule
  \end{tabular}
  
    \caption{Ablation study on loss components of our method on LEVIR-CD~\cite{chen_spatial-temporal_2020} with batch size 64, evaluated on the validation split.}
    \label{tab:ablation_loss}
\end{table}

\begin{table}[h]
\centering
\small
\renewcommand{\arraystretch}{1.05}
\setlength{\tabcolsep}{3pt}
\begin{tabular}{l c c c}
\toprule
Method 
& Trainable Params (M)
& GFLOPs
& Time (ms) \\
\midrule
M-CD  & 69.81 & 28.23 & 33.84 \\
Ours  & 392.00 & 159.54 & 36.79 \\
\bottomrule
\end{tabular}
\caption{
Efficiency comparison between M-CD~\cite{paranjape_mamba-based_2024} and our proposed method. 
Inference time is measured per image pair on an RTX 5090 GPU.
}
\label{tab:efficiency_comparison}
\end{table}

\subsection{Results}
We first compare NeXt2Former-CD against recent state-of-the-art methods. Table~\ref{tab:mcd_sota_comparison} shows that our model achieves the best overall performance across LEVIR-CD~\cite{chen_spatial-temporal_2020}, WHU-CD~\cite{ji_fully_2019}, and CDD~\cite{lebedev_change_2018}. In addition, Table~\ref{tab:ablation_deform_attn} indicates that deformable attention in the Feature Fusion Module consistently improves the validation metrics over standard cross-attention, supporting its use for handling residual spatial offsets and boundary-level correspondences.

Table~\ref{tab:efficiency_comparison} summarizes efficiency. Despite using a larger backbone, our inference latency remains comparable to M-CD on RTX 5090, reflecting the strong GPU parallelism of convolutional and attention-based components.

\begin{figure}[t]
\centering
\includegraphics[width=\columnwidth]{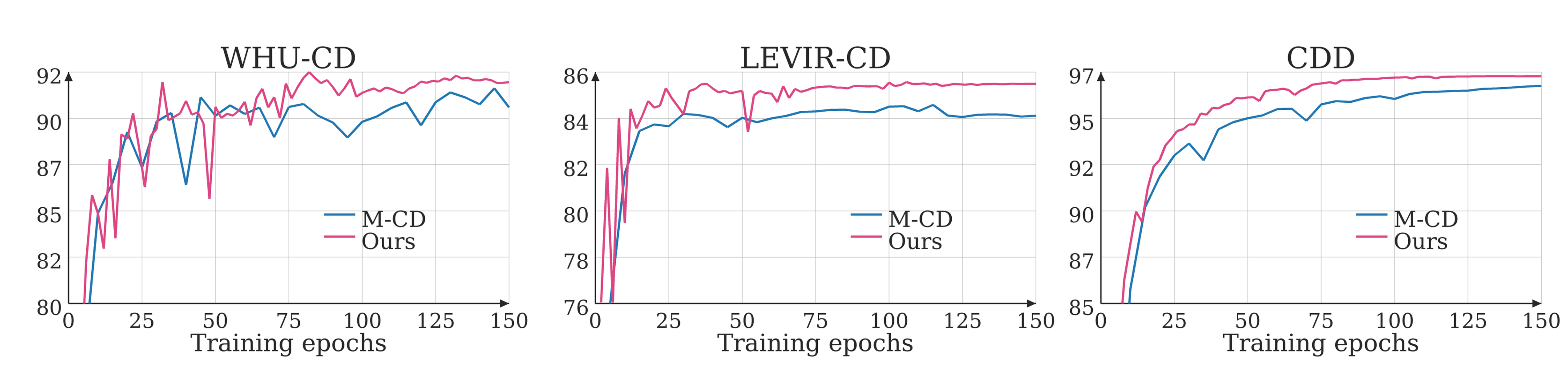}
\caption{
Validation IoU curves for M-CD~\cite{paranjape_mamba-based_2024} and the proposed method. The vertical axis shows IoU for the change class.
}
\label{fig:iou_curves}
\end{figure}

Training dynamics are further illustrated in Fig.~\ref{fig:iou_curves}. The validation IoU curves indicate that our method reaches strong performance earlier than the M-CD baseline across all three datasets. On WHU-CD and CDD, NeXt2Former-CD attains a high IoU level within the first $\sim$25 epochs and then continues to improve gradually with minor fluctuations. We attribute this faster early-stage progress to the robust representations provided by the DINOv3 pre-trained encoder, which offers a strong initialization for downstream change detection.

To better understand these numerical gains, we visualize the change masks in Fig.~\ref{fig:qualitative}. In the first row (WHU-CD), the baseline M-CD predicts masks with jagged boundaries for large building structures, whereas our model produces edges that align more closely with the ground truth. This improvement suggests that our decoder design is effective at maintaining shape integrity. The third and fourth rows (LEVIR-CD) show that the baseline model is prone to false positives in unchanged background areas, likely due to seasonal variations. Our approach suppresses these nuisance factors more effectively. Finally, in the complex scenes of the CDD dataset (last row), our model detects the changed object more completely where the baseline misses significant portions. This observation aligns with our ablation results, supporting the conclusion that deformable attention helps in aggregating context for targets with complex spatial variations.

\section{Conclusion}

In this paper, we presented NeXt2Former-CD, a unified framework for bi-temporal change detection. By combining a DINOv3-pretrained ConvNeXt backbone with deformable attention fusion and a Mask2Former decoder, the proposed pipeline supports multi-scale temporal interaction and is intended to better tolerate spatial displacement in bi-temporal imagery. Experiments on three standard benchmarks show improved F1/IoU (and OA) compared with recent Mamba-based baselines under the same training protocol. Additionally, our computational analysis reveals that the proposed architecture achieves a favorable trade-off between performance and efficiency, maintaining competitive inference speeds on parallel hardware. Overall, these results provide evidence that well-optimized 2D convolutional and Transformer-style components remain highly competitive for change detection, and motivate a broader re-examination of architecture choices beyond SSM-centric designs for future high-resolution remote sensing systems.

\bibliographystyle{IEEEbib}
\bibliography{strings,refs,ICIP2026_ChangeDetection}

\setcounter{section}{0}
\setcounter{figure}{0}
\setcounter{table}{0}
\renewcommand{\thesection}{S\arabic{section}}
\renewcommand{\thesubsection}{S\arabic{section}.\arabic{subsection}}
\renewcommand{\thefigure}{S\arabic{figure}}
\renewcommand{\thetable}{S\arabic{table}}

\section*{Supplementary Material}


\section{F1 Re-evaluation for M-CD}

During our comparative analysis, we revisited the evaluation procedure of M-CD~\cite{paranjape_mamba-based_2024}.
We observed that the official evaluation code does not directly compute the F1 score,
but instead derives it indirectly from other reported metrics.

To verify the consistency of the reported results, we re-evaluated the officially
released M-CD~\cite{paranjape_mamba-based_2024} checkpoints using the provided model weights.
We confirmed that the Intersection-over-Union (IoU) and Overall Pixel Accuracy (OA)
match the values reported in the original paper.

To ensure metric consistency, we modified the evaluation code to compute the F1 score
directly during evaluation rather than interpolating it from other metrics.
This modification follows the standard definition of F1 used in binary semantic segmentation.

Change detection can be formulated as a binary segmentation problem, where F1 and IoU
for the change class are mathematically related:

\begin{equation}
F1 = \frac{2 \cdot IoU}{1 + IoU}
\end{equation}

For example, the original M-CD~\cite{paranjape_mamba-based_2024} paper reports an F1 score of 0.921 and an IoU of 0.850
on the LEVIR-CD~\cite{chen_spatial-temporal_2020} dataset.
Since IoU is reported to three decimal places, the underlying IoU is in
$[0.8495,\,0.8505)$.
At the upper bound IoU=$0.8505$, the corresponding F1 is $0.9192$,
which is still below the reported F1=$0.921$.

This discrepancy suggests that the reported F1 score is not fully consistent with
the stated IoU value under the standard metric definition.

Therefore, for a fair comparison, we recomputed the F1 scores by directly evaluating
the released M-CD~\cite{paranjape_mamba-based_2024} checkpoints using a consistent evaluation protocol.
Table~\ref{tab:mcd_f1_reval} summarizes the F1 scores reported in the original paper
and those obtained through our re-evaluation.

\begin{table}[h]
  \centering
  \renewcommand{\arraystretch}{1.15}
  \setlength{\tabcolsep}{3pt}
  \begin{tabular}{l c c c}
    \toprule
    & WHU-CD & LEVIR-CD & CDD \\
    \midrule
    Reported in M-CD
      & 0.953 & 0.921 & 0.982 \\
    Re-evaluated 
      & 0.954 & 0.919 & 0.981 \\
    \bottomrule
  \end{tabular}
  \caption{F1 scores reported in M-CD~\cite{paranjape_mamba-based_2024} and those obtained by re-evaluating the officially released checkpoints.}
  \label{tab:mcd_f1_reval}
\end{table}

\section{Encoder Backbone Selection}

DINOv3~\cite{simeoni_dinov3_2025} offers a range of pretrained visual encoders covering both Vision Transformer (ViT) and ConvNeXt architectures, trained on different large-scale datasets. In particular, ViT models are available with pretraining on either the web-scale LVD-1689M dataset or the satellite-domain SAT-493M dataset, while ConvNeXt models are pretrained on LVD-1689M.

We first examine the impact of encoder architecture and pretraining data. As reported in Table~\ref{tab:ablation_pretrain_data}, ConvNeXt-L pretrained on LVD-1689M achieves substantially better performance than the ViT-based alternatives. ConvNeXt-L reaches an IoU of 0.8515 on LEVIR-CD~\cite{chen_spatial-temporal_2020}, compared to 0.8300 for ViT-L/16 (SAT-493M) and 0.8330 for ViT-H+/16 (LVD-1689M). These results indicate that, within our framework, ConvNeXt encoders provide more suitable representations for change detection than ViT encoders, even when ViT models are pretrained on satellite-domain data.

\begin{table}[h]
  \centering
  \renewcommand{\arraystretch}{1.15}
  \setlength{\tabcolsep}{3pt}
  \begin{tabular}{p{0.26\columnwidth} c c c}
    \toprule
    Backbone
    & \#Params
    & Pretrain
    & LEVIR-CD IoU $\uparrow$ \\

    \midrule
    ViT-L/16     & 300M & SAT & 0.8300 \\

    \midrule
    ViT-H+/16    & 840M & LVD & 0.8330 \\
    
    \midrule
    ConvNeXt-L         & 198M & LVD & \textbf{0.8515} \\

    \bottomrule
  \end{tabular}
    \caption{
    Effect of encoder backbone and pretraining data for DINOv3~\cite{simeoni_dinov3_2025} models on LEVIR-CD~\cite{chen_spatial-temporal_2020}.
    All models are trained with batch size 64 using DINOv3 encoders, the original FRM and FFM with cross-attention,
    and a Mask2Former~\cite{cheng_masked-attention_2022} decoder with the original set-based loss.
    LVD denotes LVD-1689M (web-scale), and SAT denotes SAT-493M (satellite-domain) pretraining.
    Results are reported on the validation split.
    }
    
  \label{tab:ablation_pretrain_data}
\end{table}

\enlargethispage{2\baselineskip}

\begin{table}[b]
  \centering
  \renewcommand{\arraystretch}{1.15}
  \setlength{\tabcolsep}{4pt}
  \begin{tabular}{p{0.34\columnwidth} c c}
    \toprule
    Backbone 
    & \#Params $\downarrow$
    & WHU-CD IoU $\uparrow$ \\
    \midrule
    ConvNeXt-Tiny  & 29M & 0.8966 \\
    ConvNeXt-Small & 50M & 0.9042 \\
    ConvNeXt-Base  & 89M & 0.9048 \\
    ConvNeXt-Large & 198M & \textbf{0.9193} \\
    \bottomrule
  \end{tabular}
  \caption{
    Effect of encoder backbone capacity for DINOv3-pretrained ConvNeXt models on WHU-CD~\cite{ji_fully_2019}.
    All models use DINOv3 encoders with identical FRM and FFM configurations, 
    a Mask2Former~\cite{cheng_masked-attention_2022} decoder, and the same training protocol.
    Only the ConvNeXt backbone scale (Tiny / Small / Base / Large) is varied.
    Results are reported on the validation split.
    }

  \label{tab:ablation_backbone_size}
\end{table}

Having selected ConvNeXt as the encoder family, we further evaluate the effect of backbone capacity. Table~\ref{tab:ablation_backbone_size} shows a consistent improvement as model scale increases. ConvNeXt-Large achieves the highest IoU of 0.9193 on WHU-CD~\cite{ji_fully_2019}, outperforming ConvNeXt-Base (0.9048), ConvNeXt-Small (0.9042), and ConvNeXt-Tiny (0.8966). This trend suggests that increased encoder capacity benefits change detection performance in our setting.

Based on these observations, we adopt the DINOv3-pretrained ConvNeXt-Large encoder in the main experiments.

\end{document}